\documentclass{article}

\pdfoutput=1
\usepackage{microtype}
\usepackage{graphicx}
\usepackage{subfigure}
\usepackage{booktabs} 
\usepackage{graphics}
\usepackage{adjustbox}
\usepackage{array}
\usepackage{multirow}
\usepackage[font=small,skip=0pt]{caption}

\usepackage{amsmath}
\usepackage[linesnumbered,ruled,vlined]{algorithm2e}

\SetCommentSty{mycommfont}

\usepackage{hyperref}

\usepackage{xspace}

\usepackage[accepted]{sysml2019}

\newcommand{\NAME}{AdaptivFloat\xspace}

\sysmltitlerunning{\NAME: A Floating-point based Data Type for Resilient Deep Learning Inference}

\begin{document}

\twocolumn[
\sysmltitle{\NAME: A Floating-point based Data Type for Resilient Deep Learning Inference}



\sysmlsetsymbol{equal}{*}

\begin{sysmlauthorlist}
\sysmlauthor{Thierry Tambe}{to}
\sysmlauthor{En-Yu Yang}{to}
\sysmlauthor{Zishen Wan}{to}
\sysmlauthor{Yuntian Deng}{to}

\sysmlauthor{Vijay Janapa Reddi}{to}
\sysmlauthor{Alexander Rush}{go}
\sysmlauthor{David Brooks}{to}
\sysmlauthor{Gu-Yeon Wei}{to}

\end{sysmlauthorlist}

\sysmlaffiliation{to}{Harvard University, Cambridge, MA, USA}

\sysmlaffiliation{go}{Cornell Tech, New York, NY, USA}


\sysmlkeywords{Machine Learning, SysML}

\vskip 0.2in

\begin{abstract}
Conventional hardware-friendly quantization methods, such as fixed-point or integer, tend to perform poorly at very low word sizes as their shrinking dynamic ranges cannot adequately capture the wide data distributions commonly seen in sequence transduction models. We present \NAME, a floating-point inspired number representation format for deep learning that dynamically maximizes and optimally clips its available dynamic range, at a layer granularity, in order to create faithful encoding of neural network parameters. \NAME consistently produces higher inference accuracies compared to block floating-point, uniform, IEEE-like float or posit encodings at very low precision ($\leq$ 8-bit) across a diverse set of state-of-the-art neural network topologies. And notably, \NAME is seen surpassing baseline FP32 performance by up to +0.3 in BLEU score and -0.75 in word error rate at weight bit widths that are $\leq$ 8-bit. Experimental results on a deep neural network (DNN) hardware accelerator, exploiting \NAME logic in its computational datapath, demonstrate per-operation energy and area that is 0.9$\times$ and 1.14$\times$, respectively, that of equivalent bit width integer-based accelerator variants. 

\end{abstract}
]



\printAffiliationsAndNotice{}  

\section{Introduction}
\label{submission}
Deep learning approaches have transformed representation learning in a multitude of tasks. Recurrent Neural Networks (RNNs) are now the standard solution for speech recognition, exhibiting remarkably low word error rates~\cite{sotaspeech} while neural machine translation has narrowed the performance gap versus human translators~\cite{googlenmt}. Convolutional Neural Networks (CNNs) are now the dominant engine behind image processing and have been pushing the frontiers in many computer vision applications~\cite{CNN12, resnet}. Today, deep neural networks (DNNs) are deployed at all computing scales, from resource-constrained IoT edge devices to massive data center farms. In order to exact higher compute density and energy efficiency on these compute platforms, a plethora of reduced precision quantization techniques have been proposed. 

\begin{figure}[!t]
    \centering
    \includegraphics[width=\linewidth]{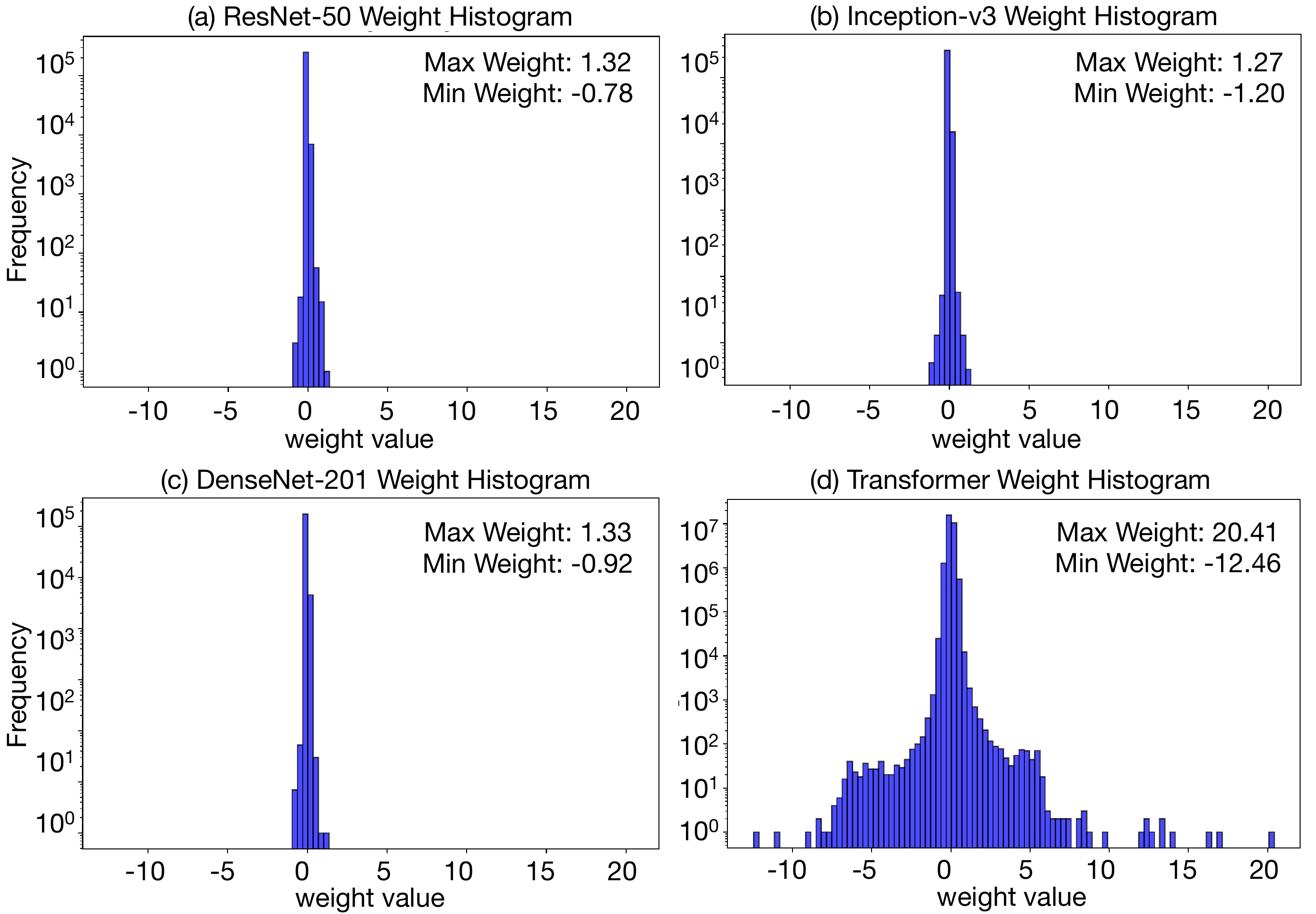} 
    \caption{Histogram of the weight distribution for (a) ResNet-50, (b) Inception-v3, (c) DenseNet-201 and (d) Transformer whose weight values are more than 10$\times$ higher than the maximum weight value from popular CNNs.}
    \label{fig:weight_distribution}
    \vspace{-1.5em}
\end{figure}

In this line of research, a large body of work has focused on fixed-point encodings~\cite{Choi2019ACCURATEAE,ibm2015,qualcomm_fixed_point,sips2014,CourbariauxBD15} or uniform quantization via integer~\cite{migacz2017, googleint}. 
These fixed-point techniques are frequently evaluated on shallow models or on CNNs exhibiting relatively narrow weight distributions. 
However, as it can be seen from Figure~\ref{fig:weight_distribution}, sequence transduction models with layer normalization such as the Transformer~\cite{attention} can contain weights more than an order of magnitude larger than those from popular CNN models with batch normalization such as ResNet-50, Inception-v3 or DenseNet-201. The reason for this phenomenon is that batch normalization effectively produces a weight normalization side effect~\cite{weightnorm} whereas layer normalization adopts invariance properties that do not reparameterize the network~\cite{layernorm}. 

In the pursuit of wider dynamic range and improved numerical accuracy, there has been surging interest in floating-point based~\cite{bfp2018, flexpoint}, logarithmic~\cite{rethinkfloat,logcnn} and posit representations~\cite{posit2017}, which also form the inspiration of this work. 

\NAME improves on the aforementioned techniques by dynamically maximizing its available dynamic range at a neural network layer granularity. And unlike block floating-point (BFP) approaches with shared exponents that may lead to degraded rendering of smaller magnitude weights, \NAME achieves higher inference accuracy by remaining committed to the standard floating-point delineation of independent exponent and mantissa bits for each tensor element. However, we break from IEEE 754 standard compliance with a unique clamping strategy for denormal numbers and with a customized proposition for zero assignment, which enables us to engineer leaner hardware.

Rather than proposing binary or ternary quantization techniques evaluated on a small number of carefully selected models, through \NAME, we aim to inform a generalized floating-point based mathematical blueprint for adaptive and resilient DNN quantization that can be easily applied on neural models of various categories (CNN, RNN or MLP), layer depths and parameter statistics. 

By virtue of an algorithm-hardware co-design, we also propose a processing element implementation that exploits the \NAME arithmetic in its computational datapath in order to yield energy efficiencies that surpass those of integer-based variants. Furthermore, owing to the superior performance of \NAME at very low word sizes, as it will be shown, higher compute density can be acquired at a lower penalty for computational accuracy compared to block floating-point, integer, or non-adaptive IEEE-like float or posit encodings. Altogether, the \NAME algorithm-hardware co-design framework offers a compelling alternative to integer or fixed-point solutions.

Finally, we note that the \NAME encoding scheme is self-supervised as it only relies on unlabeled data distributions in the network. 


This paper makes the following contributions:
\begin{itemize} 

\item We propose and describe \NAME: a floating-point based data encoding algorithm for deep learning, which maximizes its dynamic range at a neural network layer granularity by dynamically shifting its exponent range and by optimally clipping its representable datapoints. 

\item We evaluate \NAME across a diverse set of DNN models and tasks and show that it achieves higher classification and prediction accuracies compared to equivalent bit width uniform, block floating-point and non-adaptive posit and float quantization techniques.

\item We propose a hybrid float-integer (HFINT) PE implementation that exploits the \NAME mechanism and provides a cost-effective compromise between the high accuracy of floating-point computations and the greater hardware density of fixed-point post-processing. We show that the HFINT PE produces higher energy efficiencies compared to conventional monolithic integer-based PEs.

\item We design and characterize an accelerator system targeted for sequence-to-sequence neural networks and show that, when integrated with HFINT PEs, lower overall power consumption compared to an integer-based adaptation is obtained.

\end{itemize}

The rest of the paper is structured as follows. A summary of prominent number and quantization schemes used in deep learning is narrated in Section~\ref{sec:related}. We present the intuition and a detailed description of the \NAME algorithm in Section~\ref{sec:meth}. The efficacy and resiliency of \NAME is demonstrated in Section~\ref{sec:experiments} across DNN models of varying parameter distributions. Section~\ref{sec:arch} describes the hardware modeling with energy, area and performance efficiency results reported in Section~\ref{sec:hardware_eval}. Section~\ref{sec:conclusion} concludes the paper.
\section{Related Work}\label{sec:related}

\textbf{Quantization Techniques.} Low-precision DNN training and inference have been researched heavily in recent years with the aim of saving energy and memory costs. A rather significant percentage of prior work in this domain~\cite{conv_mobiles,wrpn,wen,dorefanet,halfwave,lqnets, deepcompression} have focused on or evaluated their low precision strategies strictly on CNNs or on models with narrow parameter distributions. 
Notably, inference performance with modest accuracy degradation has been demonstrated with binary~\cite{binarynet}, ternary~\cite{ternary}, and quaternary weight precision~\cite{Choi2019ACCURATEAE}. Often, tricks such as skipping quantization on the sensitive first and last layers are performed in order to escape steeper end-to-end accuracy loss.

Extending these aggressive quantization techniques to RNNs have been reported~\cite{qrnn}, although still with recurrent models exhibiting the same narrow distribution seen in many CNNs. \cite{Park2018} noticed that large magnitude weights bear a higher impact on model performance and proposed outlier-aware quantization, which requires separate low and high bit-width precision for small and outlier weight values, respectively. However, this technique complicates the hardware implementation by requiring two separate PE datapaths for the small and the outlier weights.

\textbf{Hardware-Friendly Encodings.} Linear fixed-point or uniform integer quantization is commonly used for deep learning hardware acceleration~\cite{Jouppi2017, googleint, minerva} as it presents an area and energy cost-effective solution compared to floating-point based processors. Moreover, low-precision integer inference has already made its way into commercial systems. NVIDIA’s TensorRT~\cite{migacz2017} is a commercial library that determines 8-bit integer quantization parameters offline for GPU inference. Google's TPU, based on INT8 computations, has been deployed in datacenters to perform accelerated inference of DNN applications. While integer quantization has been favorably applied to CNNs, ~\cite{int8_transformer} demonstrated robust INT8 quantization on the Transformer network. In this paper, a broader quantization study on the same network is given with \NAME showing minor degradation when the weight size is as low as 5-bit. 

Aware of the dynamic range limitation of fixed-point encoding, we have seen block floating-point data types, such as Flexpoint~\cite{flexpoint} and inspired variants employed in the Brainwave NPU~\cite{brainwave}. 
Block floating-point's appeal stems from its potential to achieve floating-point-like dynamic range with hardware cost and implementation comparable to fixed-point. However, by collapsing the exponent value of each tensor element to that of the element with the highest magnitude, elements with smaller magnitudes will be more prone to data loss. Logarithmic approaches~\cite{logbase,lognet} replacing fixed-point multipliers with shifters have demonstrated smaller power consumption and higher throughput compared to linear-based architectures.

\textbf{Number Formats with Higher Dynamic Range.} For DNN computation workloads with very large operands, 16-bit number formats such as INT16 and FP16 have been adopted. And several commercial hardware accelerators such as the 2$^{nd}$ generation TPU and Intel FPGAs have opted to use the Bfloat16 data type as it preserves the dynamic range of 32-bit float by retaining its eight exponent bits, but incurs reduced precision with 7 fractional bits. 

Additionally, there has been increasing interest in using the posit data type in deep learning due to its ability to exact higher accuracy and larger dynamic range compared to floats~\cite{posit2017}. In particular, posit's tapered precision can represent small values, where the majority of DNN parameters tend to fall, more accurately than floating-point numbers. However, although posit offers a wider reach compared to floats, its accuracy on larger numbers, for a given bit width, may be lower than floats. Furthermore, due to the dynamic nature of its regime bits, hardware implementation of a posit-based datapath exhibits worse energy-delay product compared to fixed-point and floating-point implementations~\cite{positron}. We include the posit numerical format in our experimental DNN performance evaluation and comparison with the \NAME data type.

\begin{figure}[!t]
    \centering
    \includegraphics[width=\linewidth]{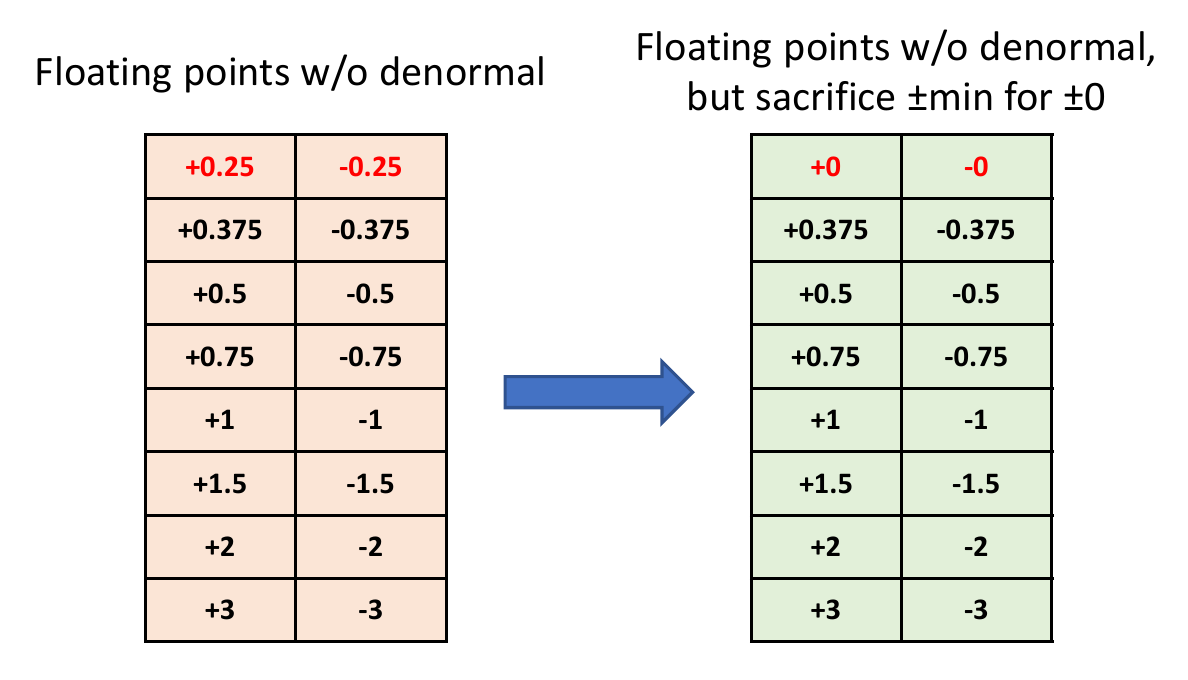}
    \caption{Illustration of the $0$ value representation in \NAME}
    \label{fig:adpfloat_zero}
    \vspace{0.5em}
\end{figure}
\section{Methodology} \label{sec:meth}
In this section, we provide greater details on \NAME, the floating-point based number encoding format for adaptive and resilient DNN quantization. We describe the inner workings behind the adjustment of the available dynamic range of representable values in order to best fit the weights of the neural network layers. 

\subsection{The \NAME Format} \label{sec:format}
    The \NAME number representation scheme generally follows the IEEE 754 Standard floating-point format that includes a sign bit, exponent bit, and mantissa bit fields. 
    In order to efficiently encode in hardware all the representable datapoints, we avoid the computation of denormal values in the \NAME number system. 
    
    The main problem arising as a result of not using denormals in floating-point is the lack of representation for the "zero" point, which is essential to neural network computations. 
    We solve this constraint by sacrificing the positive and negative minimum values to allocate the slot for "zero" as shown in Figure \ref{fig:adpfloat_zero}.
    
    Moreover, at very low bit compression, customization based on the value range of a neural network layer can greatly reduce the quantization error with little overhead on the shared extra parameters. 
    Thus, similar to integer quantization that uses a quantization scale (or step), we introduce a bias value, $exp_{bias}$, to dynamically shift the range of exponent values at a layer granularity. The calculation of $exp_{bias}$ is described in Section~\ref{sec:quantization}. A benefit of using $exp_{bias}$ is the simplicity of the hardware logic required to perform the adaptive operation compared to the multiplying quantization scale used in integer quantization. This contrast is discussed in detail in Section~\ref{sec:arch}.
    
  Algorithm \ref{alg:bit2value} shows how a bit vector, with a $exp_{bias}$ generated per layer, is converted to its decimal representation. If both exponent bits and mantissa bits are zeros, the bit vector should be interpreted as zero. Otherwise, the bit vector is converted by the following equation:
    $$ \mbox{sign}*2^{ \mbox{ exponent\ value } }* \mbox{ mantissa\ value } $$
    where the {\it exponent value} is the addition of exponent bits and $exp_{bias}$, and the {\it mantissa value} is calculated by appending an implied ``one'' as the MSB and follow the same format as standard floating point.

\begin{algorithm}[tb]
    \caption{\NAME Bit Vector to Value}
    \label{alg:bit2value}
\begin{algorithmic}
    \STATE {\bfseries Input:} bit vector $x$, number of bits $n$, number of exponent bits $e$ and $exp_{bias}$

    \vspace{0.25cm}
    \tcp{Get Mantissa bits}
    \STATE $m := n - e - 1$    

    \vspace{0.25cm}
    \tcp{Extract sign, exponent, mantissa}
    \STATE $sign := 1$ if $x[n-1] = 0$, otherwise $-1$
    \STATE $exp := x[n-2:m] + exp_{bias}$
    \STATE $mant := 1 + x[m-1:0]/2^m$

    \vspace{0.25cm}
    \tcp{Map to 0 if exp, mant bits are zeros}
    \IF {$x[n-2:0] = 0$} 
    \STATE $val := 0$    
    \ELSE
    \STATE $val := sign*2^{exp}*mant$
    \ENDIF
    \STATE {\bfseries return} $val$
\end{algorithmic}
\end{algorithm}
    
\subsection{Quantization} \label{sec:quantization}
    Having defined the \NAME data format, the quantization problem is simply mapping a full precision value to the nearest representable datapoint. 
    For this purpose, we need to determine the optimal \NAME $exp_{bias}$ which will maximize the available dynamic range of the encoding format in order to provide the most accurate rendering of a particular matrix (or NN layer). This is analogous to determining the quantization scale for integer quantizations, but $exp_{bias}$ is a small, typically negative, integer value rather than the high-precision floating-point scaling factor needed in integer quantization~\cite{migacz2017}. 
    
    Algorithm \ref{alg:quantize_adpfloat} describes how to find the most suitable $exp_{bias}$ in order to encode, as faithfully as possible, the \NAME-quantized weight matrix. We first compute the sign matrix, $W_{sign}$, and the matrix of absolute values, $W_{abs}$, from the full precision weight matrix $W_{fp}$. Then, the algorithm finds the maximum absolute value from $W_{abs}$ to determine the $exp_{bias}$ corresponding to a suitable range of representable datapoints for the weight matrix to quantize. Before doing quantization on $W_{abs}$, we first round the values smaller than the \NAME minimum representable value to zero or $value_{min}$ at a halfway threshold. Then, we clamp values larger than the max value, $value_{max}$, to $value_{max}$. 
    The quantization involves rewriting $W_{abs}$ into normalized exponent and mantissa form with an exponent matrix $W_{exp}$ and a mantissa matrix $W_{mant}$. The mantissa matrix is quantized by the quantization scale calculated by the number of mantissa bits. $W_q$ indicates the quantized mantissa matrix. Finally, the \NAME-quantized matrix is reconstructed by multiplication of $W_{sign}, 2^{W_{exp}}$, and $W_{q}$.
    
\begin{figure*}[t]
    \centering
    \includegraphics[width=0.99\linewidth]{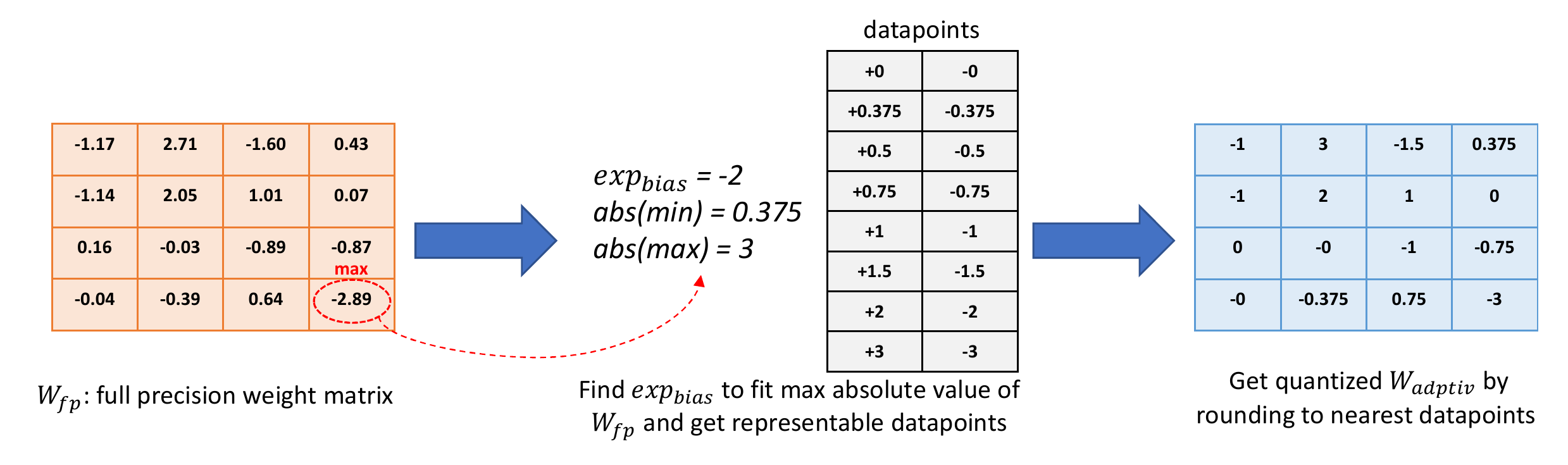}
    \caption{Illustration of $\NAME<4,2>$ quantization from a full precision weight matrix}
    \label{fig:quantize_overiew}
    \vspace{-0em}
\end{figure*}

    
    We use the notation $\NAME<n,e>$ to indicate a $n$-bit \NAME number with $e$ exponent bits. Figure \ref{fig:quantize_overiew} provides an illustration showing how the $exp_{bias}$ is chosen to best fit the range of values in a weight matrix and the resulting quantized datapoints adhering to the $\NAME<4,2>$ format. The bit vectors containing the \NAME quantized values can be packed and stored in hardware memory resources.
    
\begin{algorithm}[tb]
    \caption{\NAME Quantization}
    \label{alg:quantize_adpfloat}
\begin{algorithmic}
    \STATE {\bfseries Input:} Matrix $W_{fp}$, number of bits $n$ and number of exponent bits $e$
    
    \tcp{Get Mantissa bits}
    \STATE $m := n - e - 1$

    \vspace{0.25cm}
    \tcp{Obtain sign and abs matrices}
    \STATE $W_{sign} := sign(W_{fp})$ 
    \STATE $W_{abs} := abs(W_{fp})$
    
    \vspace{0.25cm}
    \tcp{Determine $exp_{bias}$ and range}
    \STATE Find normalized $exp_{max}$ for $max(W_{abs})$ such that 
        \\\hspace{0.5cm} $2^{exp_{max}} \leq max(W_{abs}) <  2^{exp_{max}+1}$
    \STATE $exp_{bias} := exp_{max} - (2^{e}-1)$
    \STATE $value_{min} := 2^{exp_{bias}}*(1+2^{-m})$ 
    \STATE $value_{max} := 2^{exp_{max}}*(2-2^{-m})$ 

    \vspace{0.25cm}
    \tcp{Handle unrepresentable values} 
    \STATE Round $value < value_{min}$ in $W_{abs}$ to $0$ or $value_{min}$
    \STATE Clamp $value > value_{max}$ in $W_{abs}$ to $value_{max}$

    \vspace{0.25cm}
    \tcp{Quantize $W_{fp}$} 
    \STATE Find normalized $W_{exp}$ and $W_{mant}$ such that 
        \\\hspace{0.5cm} $W_{abs} = 2^{W_{exp}}*W_{mant}$, and $1 \leq W_{mant} < 2$
    \STATE $W_{q} := $ quantize and round $W_{mant}$ by $scale = 2^{-m}$ 

    \vspace{0.25cm}
    \tcp{Reconstruct output matrix}
    \STATE $W_{adptiv} := W_{sign}*2^{W_{exp}}*W_{q}$
    
    \STATE {\bfseries return} $W_{adptiv}$
\end{algorithmic}
\vspace{-0.5em}
\end{algorithm}


\begin{figure*}[t]
    \centering
    \includegraphics[width=0.99\linewidth]{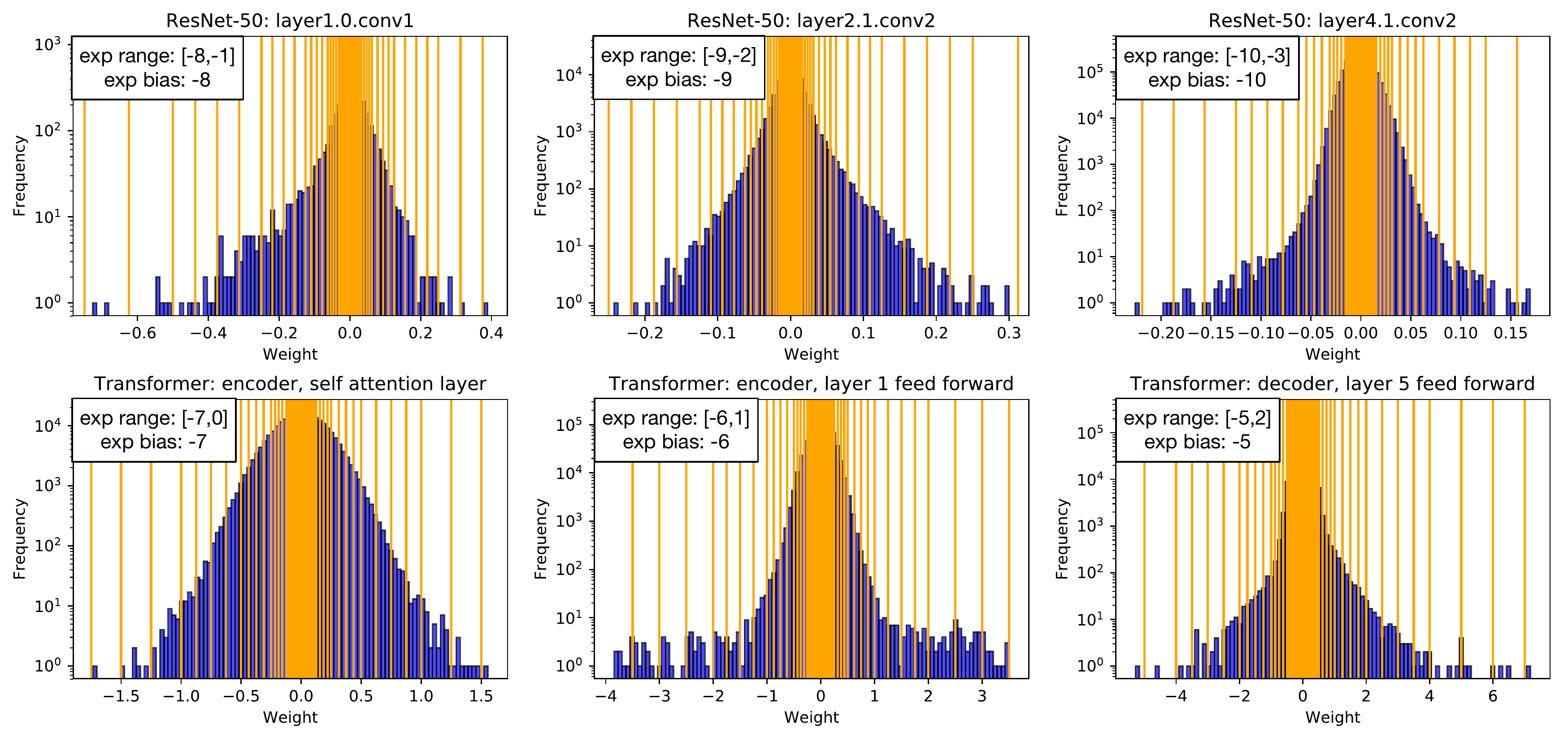}
    \caption{$\NAME<6,3>$ quantization on distinct layers of ResNet-50 (top row) and Transformer (bottom row)}
    \label{fig:adap_quant_6}
    \vspace{-1.5em} 
\end{figure*}

\subsection{Shifting the \NAME Range}
    The main characteristic of the \NAME quantization is using the $exp_{bias}$ to shift the range of quantized datapoints. Figure \ref{fig:adap_quant_6} depicts an $\NAME<6,3>$ quantization strategy applied on different ResNet-50 and Transformer layers, having varying weight distribution range. For ResNet-50, we show the convolution layers with 1x1 and 3x3 weight kernels as examples, and for the Transformer, we show two encoder layers and one decoder layer. The plots illustrate how the $exp_{bias}$ can vary from one layer to another in order to best fit their natural weight distribution. For example, the $exp_{bias}$ of ResNet-50 can be seen adjusting from -8 to -10 while $exp_{bias}$ transitions from -5 to -7 in the Transformer network. The narrower the weight distribution becomes, which indicates a smaller maximum value in the weight tensor, the more negative $exp_{bias}$ gets. 
    
\section{Experimental Results}\label{sec:experiments}

\begin{table*}[t] 
 \caption{DNN models under evaluation}
 \label{tab:model_types}
 \vskip 0.05in
  \begin{center}
 \begin{small}
 \begin{sc}
 \begin{adjustbox}{width=\textwidth, center}
\begin{tabular}{lllllll}
\toprule
Model & Application & Dataset & Structure & Number of parameters & Range of weights & FP32 performance \\
\midrule
Transformer & Machine translation & WMT'17 En-to-De & Attention, FC layers & 93M & {[}-12.46 , 20.41{]} & BLEU: 27.40  \\
Seq2Seq & Speech-to-text & LibriSpeech 960h & Attention, LSTM, FC layers & 20M & {[}-2.21 , 2.39{]} & WER: 13.34 \\
ResNet-50 & Image classification & ImageNet & CNN, FC layers & 25M & {[}-0.78, 1.32{]} & Top-1 Acc: 76.2 \\
\bottomrule
\end{tabular}
\end{adjustbox}
\end{sc}
\end{small}
\end{center}
\vskip -0.1in
\end{table*}
\begin{table}[t] 
 \caption{Impact of various weight encodings after post-training quantization (PTQ) / and after quantization-aware retraining (QAR) on the BLEU score of the Transformer model~\cite{attention} evaluated on the WMT'17 English-German dataset. FP32 performance is 27.4}
 \label{tab:transformer_weight_quant}
  \begin{center}
 \begin{small}
 \begin{sc}
 \begin{adjustbox}{width=\columnwidth,center}
\begin{tabular}{llllll}
\toprule
Bit Width & Float & BFP & Uniform & Posit & \textbf{\NAME} \\
\midrule
16 & 27.4 / 27.4  & 27.4 / 27.4  & 27.4 / 27.4 & 27.4 / 27.5  & 27.4 / 27.6\\
8  & 27.2 / 27.5  & 26.3 / 27.3  & 27.3 / 27.4 & 27.3 / 27.5 & 27.3 / 27.7\\
7  & 27.1 / 27.5  & 16.9 / 26.8  & 26.0 / 27.2 & 27.3 / 27.4  & 27.3 / 27.7 \\
6  & 26.5 / 27.1  & 0.16 / 8.4   & 0.9  / 23.5 & 26.7 / 27.2 & 27.2 / 27.6\\
5  & 24.2 / 25.6  & 0.0 / 0.0    & 0.0 / 0.0   & 25.8 / 26.6 & 26.4 / 27.3\\
4  & 0.0 / 0.0    & 0.0 / 0.0    & 0.0 / 0.0   & 0.0 / 0.0  & 16.3 / 25.5\\         
\bottomrule
\end{tabular}
\end{adjustbox}
\end{sc}
\end{small}
\end{center}
\vskip -0.1in
\end{table}

\begin{table}[t]
 \caption{Impact of weight encodings after PTQ / and after QAR on the word error rate of the LSTM and attention based seq2seq model~\cite{seq2seq} evaluated on the LibriSpeech dataset. FP32 performance is 13.34}
 \label{tab:seq2seq_weight_quant}
  \begin{center}
 \begin{small}
 \begin{sc}
 \begin{adjustbox}{width=\columnwidth,center}
\begin{tabular}{llllll}
\toprule
Bit Width & Float & BFP & Uniform & Posit & \textbf{\NAME} \\
\midrule
16 & 13.40 / 13.07  & 13.30 / 13.14  & 13.27 / 12.82 & 13.29 / 13.05  & 13.27 / 12.93\\
8  & 14.06 / 12.74  & 13.23 / 13.01  & 13.28 / 12.89 & 13.24 / 12.88 & 13.11 / 12.59\\
7  & 13.95 / 12.84  & 13.54 / 13.27  & 13.45 / 13.37 & 13.36 / 12.74  & 13.19 / 12.80 \\
6  & 15.53 / 13.48  & 14.72 / 14.74   & 14.05 / 13.90 & 15.13 / 13.88 & 13.19 / 12.93 \\
5  & 20.86 / 19.63  & 21.28 / 21.18    & 16.53 / 16.25 & 19.65 / 19.13 & 15.027 / 12.78\\
4  & inf / inf    & 76.05 / 75.65    & 44.55 / 45.99   & inf / inf  & 19.82 / 15.84\\         
\bottomrule
\end{tabular}
\end{adjustbox}
\end{sc}
\end{small}
\end{center}
\vskip -0.1in
\end{table}
\begin{table}[t] 
 \caption{Impact of weight encodings after PTQ / and after QAR on the Top-1 accuracy of ResNet-50~\cite{resnet} evaluated on the ImageNet dataset. FP32 performance is 76.2}
 \label{tab:resnet_weight_quant}
  \begin{center}
 \begin{small}
 \begin{sc}
 \begin{adjustbox}{width=\columnwidth,center}
\begin{tabular}{llllll}
\toprule
Bit Width & Float & BFP & Uniform & Posit & \textbf{\NAME} \\
\midrule
16 & 76.1 / 76.3  & 76.2 / 76.3  & 76.1 / 76.3 & 76.1 / 76.3  & 76.2 / 76.3\\
8  & 75.4 / 75.9  & 75.7 / 76.0  & 75.9 / 76.1 & 75.4 / 76.0 & 75.7 / 76.3\\
7  & 73.8 / 75.6  & 74.6 / 75.9  & 75.3 / 75.9 & 74.1 / 75.8  & 75.6 / 76.1 \\
6  & 65.7 / 74.8  & 66.9 / 74.9   & 72.9 / 75.2 & 68.8 / 75.0 & 73.9 / 75.9 \\
5  & 16.1 / 73.6  & 13.2 / 73.4    & 15.1 / 74.0 & 33.0 / 73.9 & 67.2 / 75.6\\
4  & 0.5 / 66.3    & 0.5 / 66.1    & 2.6 / 67.4  & 0.7 / 66.7  & 29.0 / 75.1\\         
\bottomrule
\end{tabular}
\end{adjustbox}
\end{sc}
\end{small}
\end{center}
\vskip -0.1in
\end{table}
For bit compression evaluation, we select three popular DNN models of distinct neural types and applications, and exhibiting relatively narrow to wide spread in their weight distributions. The models considered, as shown in Table~\ref{tab:model_types}, are: (1) Transformer~\cite{attention}, which made a very consequential impact in the field of machine translation and question answering; (2) a 4-layer LSTM encoder, 1-layer LSTM decoder, attention-based sequence-to-sequence (seq2seq) network~\cite{seq2seq} commonly used in speech recognition; and (3) ResNet-50, a well-known image classification CNN~\cite{resnet}. The Transformer and the LSTM-based Seq2Seq networks are trained on the OpenNMT platform~\cite{onmt} using the WMT'17 English-to-German and LibriSpeech datasets, respectively. And, the Pytorch toolkit~\cite{pytorch_imagenet} is used to train ResNet-50 using the ImageNet dataset.

We compare the efficacy of \NAME along with numerical data types frequently employed for deep learning acceleration, namely block floating-point (BFP), IEEE-like float, posit, and uniform representations. We created templates of these data types in Python to be run within the PyTorch framework. The \NAME, uniform, and block floating-point quantization schemes are self-adaptive in the sense that their dynamic range auto-adjusts based on the distribution of the data. The number of exponent bits in the \NAME, IEEE-like float, and posit formats is set evenly for all the layers in the network to the value yielding the highest inference accuracy after doing a search on the exponent width. Generally, the best inference performance was obtained with the exponent space set to 3 bits for \NAME, 4 bits for float (3 bits when the word size becomes 4 bits), and 1 bit for posit (0 bits when the word size becomes 4 bits). 

Finally, we note that the following results are generated by quantizing all of the layers in the DNN models in order to capture the whole-length performance of these five numerical data types unlike several works~\cite{Choi2019ACCURATEAE, dorefanet, wrpn} that intentionally skip quantization for the first and last layers.

\subsection{Root Mean Squared Error}

\begin{figure*}[t]
    \centering
    \includegraphics[width=0.99\linewidth]{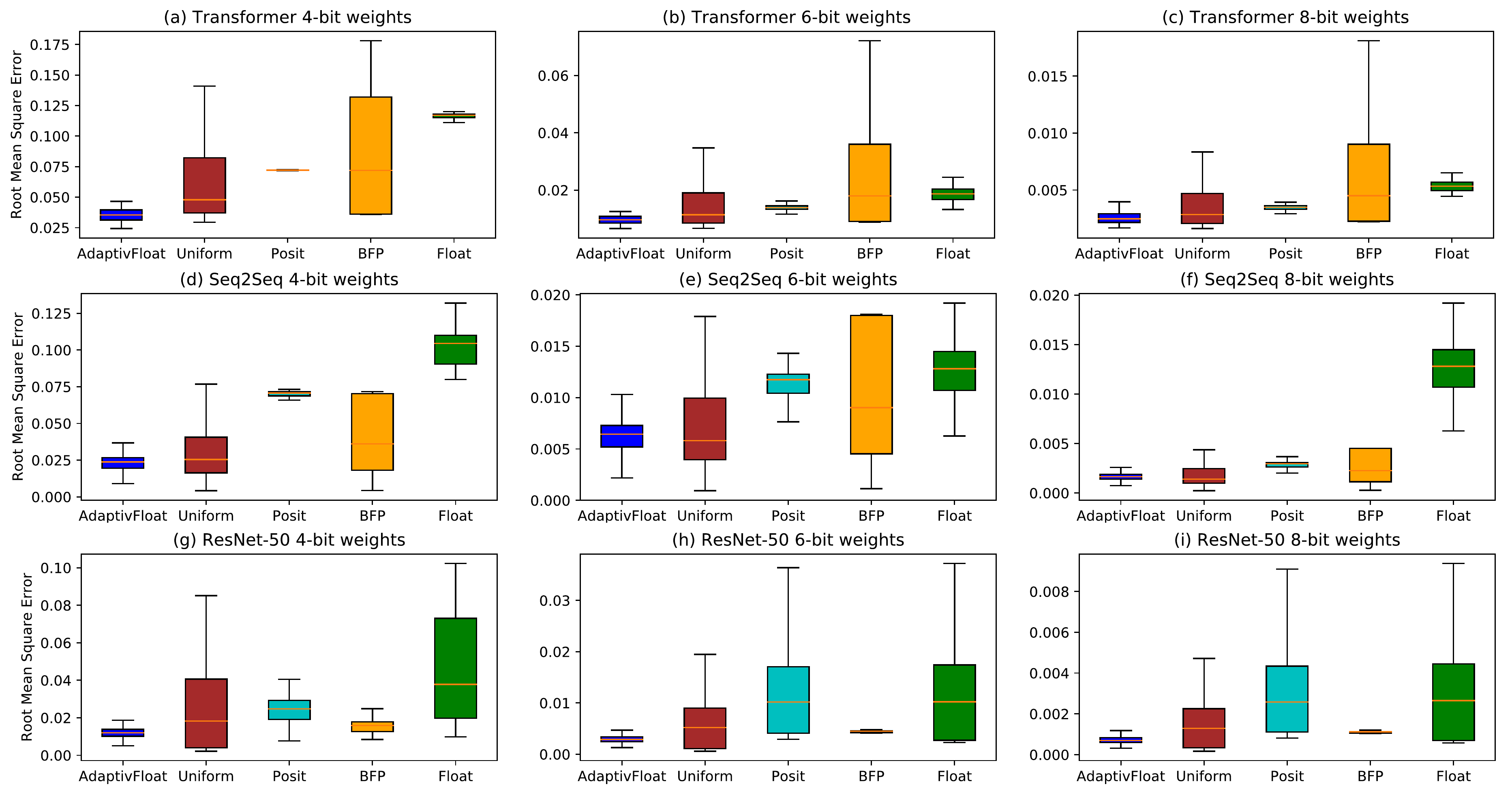}
    \caption{Root mean square of the quantization error w.r.t. FP32 at 4-bit, 6-bit and 8-bit weight precision across the layers of the Transformer, Seq2Seq and ResNet-50 models. \NAME produces the lowest mean error compared to the other number systems. 
    }
    \label{fig:rmse}
    \vspace{-1.5em} 
\end{figure*}

We begin by quantifying the quantization error of the number formats with respect to baseline FP32 precision. The boxplots depicted in Figure~\ref{fig:rmse} show the distribution of the root mean squared (RMS) quantization error emanating from the data types and computed across all layers of the three models under evaluation. \NAME consistently produces lower average quantization error compared to uniform, BFP, posit, or IEEE-like float encoding. Furthermore, among the self-adaptive data types, \NAME exhibits the tightest error spread for all bit widths of the Transformer and seq2seq networks, while BFP's error spread is thinnest for the 6-bit and 8-bit versions of the ResNet-50 model -- although with a higher mean compared to \NAME. This suggests that BFP would fare best in networks with slimmer weight distribution such as ResNet-50. Among the non-adaptive data types, we see that posit generally yields both a lower average RMS quantization error and a narrower interquartile error range compared to Float. These results provide important insights to quantized DNN performance as we dive into the bare inference accuracy results in the next subsection. 

\subsection{Inference Performance Analysis}
 Tables~\ref{tab:transformer_weight_quant}, ~\ref{tab:seq2seq_weight_quant}, and ~\ref{tab:resnet_weight_quant} show the resiliency of the data types under study as they are put to the test under varying weight bit compression on the Transformer, sequence-to-sequence, and ResNet-50 models, respectively. The inference results are tabulated after post-training quantization (PTQ) and after quantization-aware re-training (QAR) from the plateaued FP32 baseline. The training setup and the hyper-parameter recipe are kept the same for all five data types under evaluation in order to impose a fair comparison.
 
 The key observation we can distill is that \NAME demonstrates much greater resiliency at very low precision ($\leq$ 6-bit) compared to the other four data formats. For instance, at 4-bit encoding, \NAME can still yield, after retraining, a decent BLEU score of 25.5 on the Transformer model while the impact from the other four number formats is catastrophic due to insufficient dynamic range or decimal accuracy.
 \begin{table}[t] 
 \caption{Impact of both weight and activation quantization on the BLEU score of the Transformer model. Baseline FP32 performance is 27.4}
 \label{tab:transformer_weight_act_quant}
  \begin{center}
 \begin{small}
 \begin{sc}
 \begin{adjustbox}{width=\columnwidth,center}
\begin{tabular}{llllll}
\toprule
Bit Width & Float & BFP & Uniform & Posit & \textbf{\NAME} \\
\midrule
W8/A8 & 27.4  & 27.4  & 10.1 & 26.9  & 27.5\\
W6/A6  & 25.9  & 0.0  & 5.7 & 25.7 & 27.1\\
W4/A4  & 0.0  & 0.0  & 0.0 & 0.0  & 0.3 \\
\bottomrule
\end{tabular}
\end{adjustbox}
\end{sc}
\end{small}
\end{center}
\vskip -0.1in
\end{table}
\begin{table}[!htp] 
 \caption{Impact of both weight and activation quantization on the WER of the Seq2Seq model. Baseline FP32 performance is 13.34}
 \label{tab:seq2seq_weight_act_quant}
  \begin{center}
 \begin{small}
 \begin{sc}
 \begin{adjustbox}{width=\columnwidth,center}
\begin{tabular}{llllll}
\toprule
Bit Width & Float & BFP & Uniform & Posit & \textbf{\NAME} \\
\midrule
W8/A8 & 12.77  & 12.86  & 12.86 & 12.96  & 12.59\\
W6/A6  & 14.58  & 14.68  & 14.04 & 14.50 & 12.79\\
W4/A4  & inf  & 78.68  & 48.86 & inf  & 21.94 \\
\bottomrule
\end{tabular}
\end{adjustbox}
\end{sc}
\end{small}
\end{center}
\vskip -0.1in
\end{table}
\begin{table}[!htp] 
 \caption{Impact of both weight and activation quantization on the Top-1 accuracy of the ResNet-50 model. Baseline FP32 performance is 76.2}
 \label{tab:resnet_weight_act_quant}
  \begin{center}
 \begin{small}
 \begin{sc}
 \begin{adjustbox}{width=\columnwidth,center}
\begin{tabular}{llllll}
\toprule
Bit Width & Float & BFP & Uniform & Posit & \textbf{\NAME} \\
\midrule
W8/A8 & 75.7  & 75.7  & 75.9 & 75.8  & 76.0 \\
W6/A6  & 73.5  & 73.4  & 74.1 & 73.6 & 75.0 \\
W4/A4  & 63.3  & 63.0  & 64.3 & 63.0 & 72.4 \\
\bottomrule
\end{tabular}
\end{adjustbox}
\end{sc}
\end{small}
\end{center}
\vskip -0.1in
\end{table}
 We can make similar observations on the seq2seq and ResNet-50 models as \NAME show modest retrained performance degradation at 4-bit and 5-bit weight precision. Notably, only a 1.2 Top-1 accuracy drop is seen with a weight width of 4-bit. When the weights of the seq2seq model are quantized to 4-bit, the non-adaptive data types (float and Posit) are essentially unable to provide expressible transcription. This suggests that, for resilient performance at very low word size, it is critical to have a quantization scheme that can adjust its available dynamic range to represent the network's weights as faithfully as possible. \NAME's observed robustness at very low precision enables higher compute density into reconfigurable architectures at a low penalty for computational accuracy. 
 
 
Adding noise to weights when computing the parameter gradients has been shown to exact a regularization effect that can improve generalization performance~\cite{deepnoise}. This effect can be seen in all data types but is particularly pronounced in \NAME with performance exceeding FP32 by up to +0.3 in BLEU score, -0.75 in word error rate and +0.1 in Top-1 accuracy. 

\subsection{Effect of both Weight and Activation Quantization}

Tables~\ref{tab:transformer_weight_act_quant}, ~\ref{tab:seq2seq_weight_act_quant}, and ~\ref{tab:resnet_weight_act_quant} report the inference performance from reducing the word size of both weights and activations on the Transformer, Seq2Seq, and ResNet-50 models, respectively. W$n$/A$n$ signifies a quantization of $n$-bit weight and $n$-bit activation.

We observe that \NAME's 8-bit performance is as good as, if not better than, the baseline FP32 result on all three DNN models while the degradation at 6-bit is still modest. 
Interestingly, in the case of the seq2seq model, the 6-bit \NAME weight and activation quantization generates regularization effective enough to exceed the FP32 baseline. 
At 4-bit weight and activation precision, the performance degradation of \NAME is steeper on the sequence models than on ResNet-50 as many of the activations from the attention mechanisms fall outside of the available dynamic range of the number format. 

\section{PE Architecture}\label{sec:arch}
\NAME's superior bit compression ability paves the way to efficient bit packing into resource-constrained accelerators. In this section, we describe the design of a hybrid Float-Integer (HFINT) PE that exploits the \NAME logic in its computational datapath and provides an efficient compromise between the high accuracy of floating-point computations and the greater hardware density of fixed-point post-processing. We contrast the proposed PE architecture against that of a conventional integer (INT) PE. 


\begin{figure}[!t]
\centering
\includegraphics[width=0.37\textwidth]{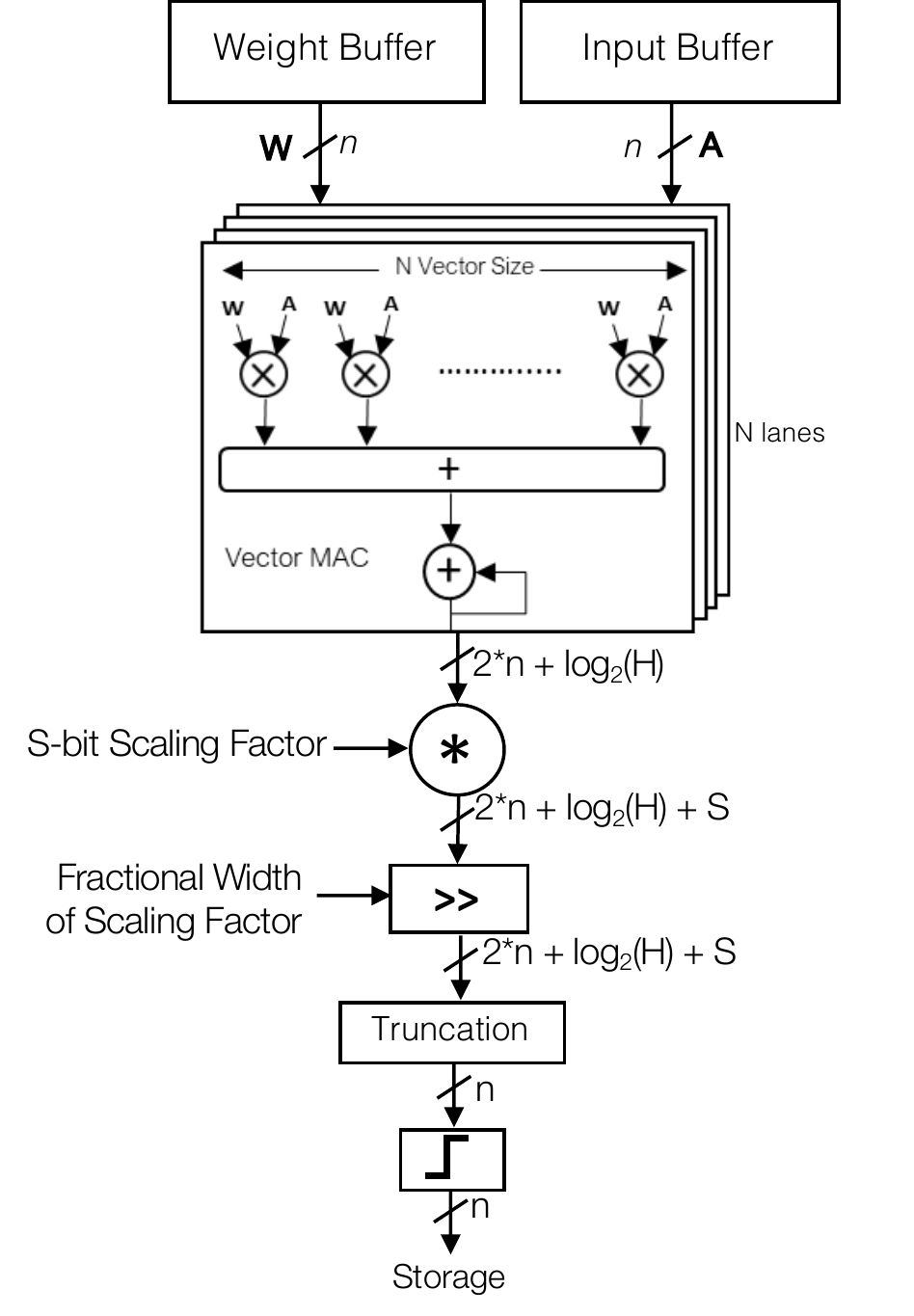} 
\caption{Conventional $n$-bit Integer-based PE.   }
\label{fig:intpe}
\vspace{-1.5em}
\end{figure}

\begin{figure}[!t]
\centering
\includegraphics[width=0.45\textwidth]{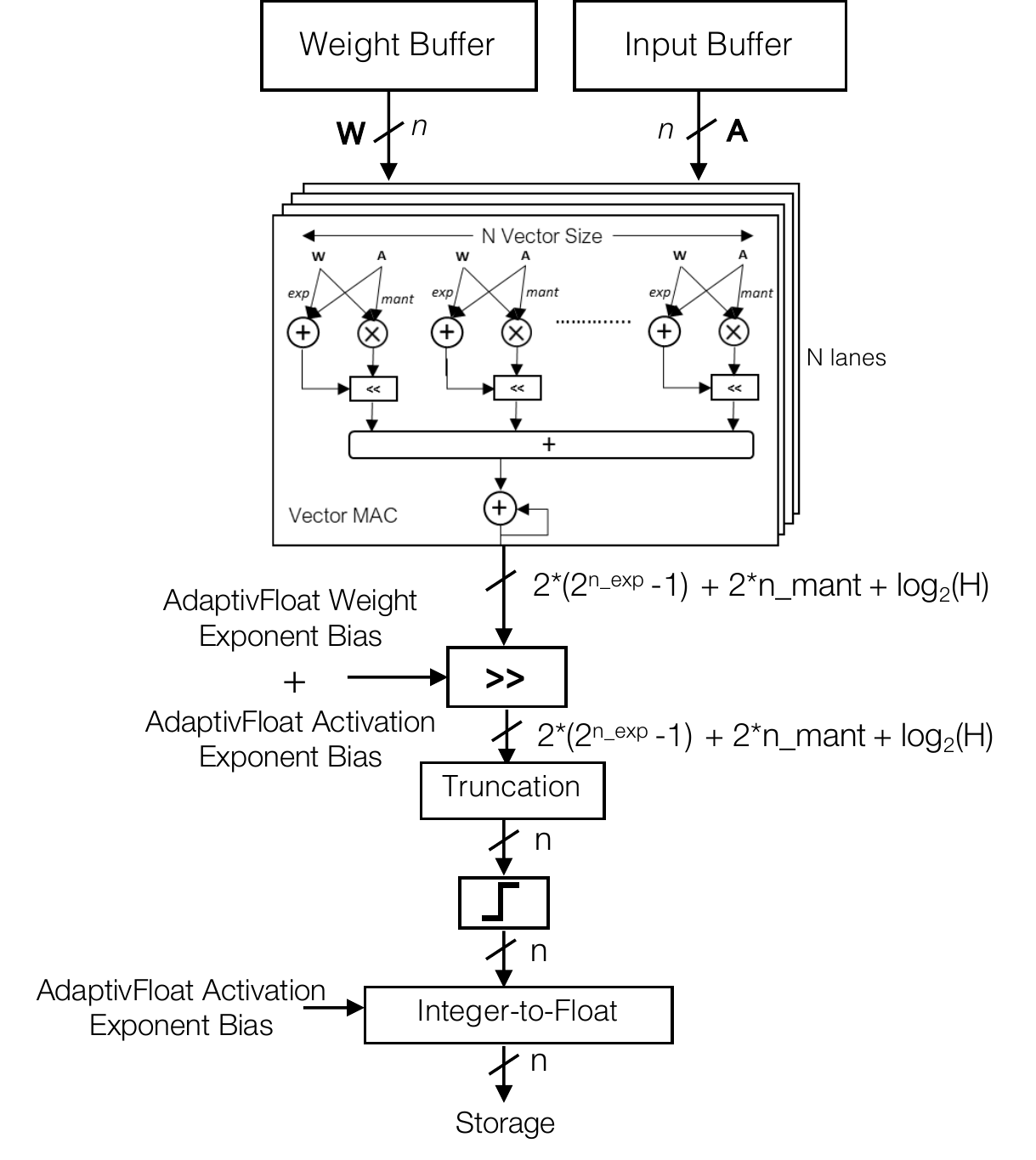} 
\caption{$n$-bit Hybrid Float-Integer PE.
  }
\label{fig:hfintpe}
\vspace{-0.5em}
\end{figure}

\subsection{Conventional Integer PE}
The micro-architecture of a $n$-bit integer-based PE is shown in Figure~\ref{fig:intpe}. It contains fixed-point vector MAC units receiving $n$-bit integer weight and activation vectors. The MAC partial sums are stored in $2*n+log_{2}(H)$-bit registers in order to accumulate up to $H$ values without overflow. A high-precision scaling factor is typically used to dequantize the computation with high accuracy~\cite{migacz2017}. Using a $S$-bit scaling factor requires the scaled results to be stored in registers of width $2*n+log_{2}(H)+S$, which later are bit-shifted right by the fractional value of the scaling. Then, the data is clipped and truncated back to $n$ bits before being modulated by the neural network activation function. An 8-bit integer-based PE architecture will be referred later in the document as INT8/24/40 to designate a datapath with 8-bit MAC operands, accumulated into 24-bit (to add up to 256 values without overflow) and then scaled to 40-bit using a 16-bit scaling factor.

\subsection{Hybrid Float-Integer PE}

\begin{figure*}[t]
\centering
\includegraphics[width=0.75\textwidth]{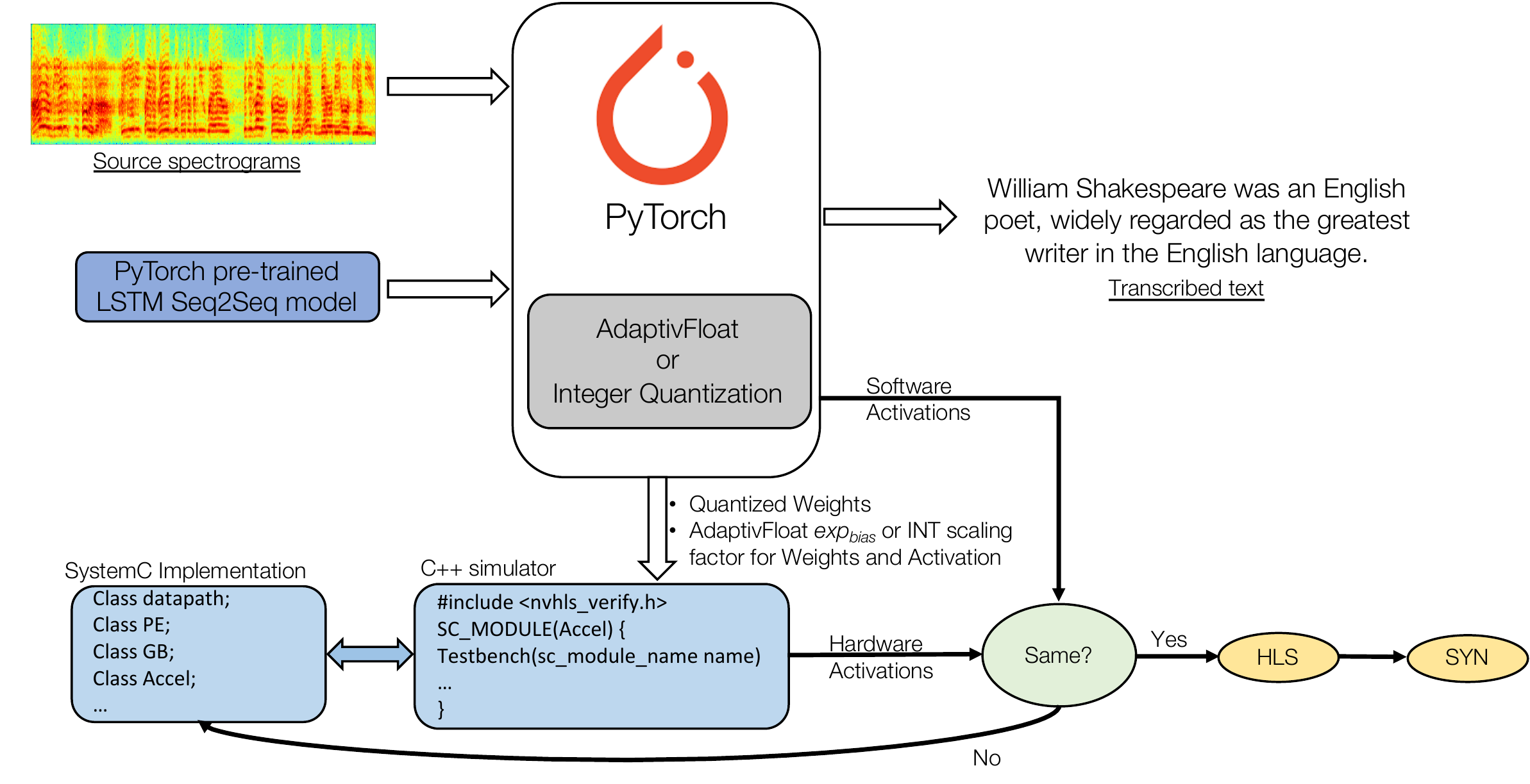} 
\caption{Algorithm-hardware co-design and verification methodology of the INT and HFINT accelerator designs
  }
\label{fig:codesign}
\end{figure*}

Figure~\ref{fig:hfintpe} illustrates the micro-architecture of a $n$-bit Hybrid Float-Integer (HFINT) PE. The vector MACs units perform floating-point multiplications between a $n$-bit float weight vector and a $n$-bit float activation vector --- and accumulate the result as integer. The weights and activations stored on-chip are quantized according to the \NAME algorithm described in Algorithm~\ref{alg:quantize_adpfloat} in Section \ref{sec:meth}. 
The extracted \NAME $exp_{bias}$ for weight and activation tensors are saved in allocated 4-bit registers and are used to shift the exponent range of the accumulated partial sums. 
We note that while the \NAME $exp_{bias}$ for the static weights are extracted post-training, the $exp_{bias}$ for the dynamic activations are informed from statistics during offline batch inference on the test dataset. The accumulation precision needs to be $2*(2^{n_{exp}} -1)+2*n_{mant}+log_{2}(H)$ -bit in order to accumulate up to $H$ values without overflow. The accumulated partial sums are then clipped and truncated back to $n$-bit integer before being processed by the activation function. At the end of the PE datapath, the integer activations are converted back to the \NAME format. The 8-bit HFINT PE architecture will be referred to as HFINT8/30 to indicate an 8-bit MAC datapath with 30-bit accumulation. 

A key contrast to note between the INT PE and the HFINT PE, apart from the differing data types employed in the vector MAC units, is the INT PE requires a post-accumulation multiplier in order to perform the adaptive operation of the quantization. This in turn increases the required post-accumulation precision by $S$-bit before truncation. In the next section, we provide energy, performance, and area comparisons between the two PE topologies. 


\section{Hardware Evaluation}\label{sec:hardware_eval}
\subsection{Algorithm-Hardware Co-design Methodology}
We developed a design and verification flow, as illustrated in Figure~\ref{fig:codesign}, which closes the loop between software modeling and backend hardware implementation. 
The \NAME and integer quantizations are performed in the Pytorch deep learning framework during and post training. 
The extracted \NAME $exp_{bias}$ for weights and activations and the scaling factor from the integer quantization are sent to the C++ simulator, along with the quantized weights. 

\begin{figure}[!t]
\centering
\includegraphics[width=0.45\textwidth]{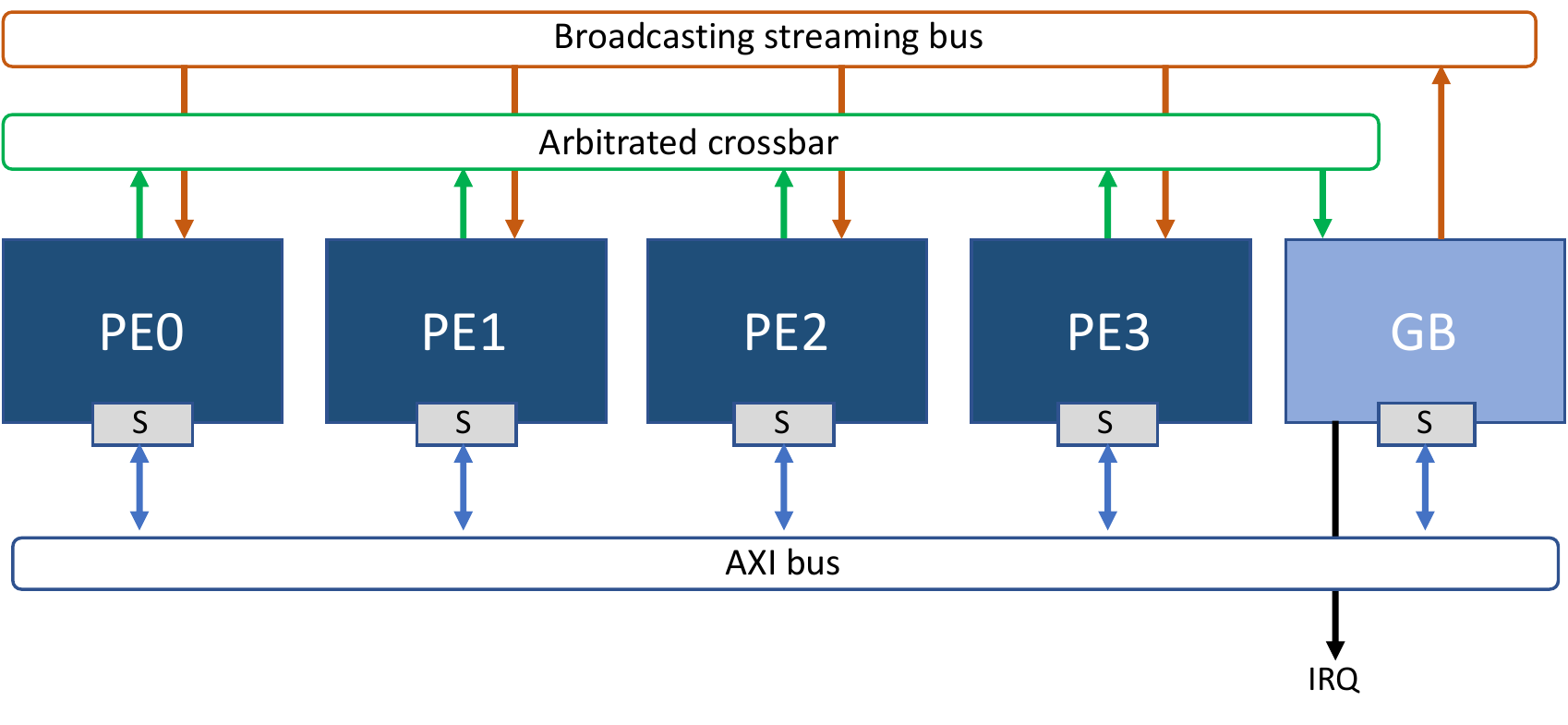} 
\caption{Accelerator system with 4 PEs and a global buffer (GB) targeting sequence-to-sequence networks.}
\label{fig:accel}
\vspace{-1.5em}
\end{figure}

In order to evaluate the hardware on a realistic DNN workload, we designed an accelerator system, depicted in Figure~\ref{fig:accel}, targeted for RNN and FC sequence-to-sequence networks where we have seen wider parameter distributions compared to convolution networks. The accelerator is evaluated with four PEs that are integrated as either INT or HFINT. 
Each PE contains an input/bias buffer with sizes ranging from 1KB to 4KB and a weight buffer whose size ranges from 256KB to 1MB depending on the vector size and operand bit width. A global buffer (GB) unit with 1MB of storage collects the computed activations from the individual PEs via the arbitrated crossbar channel and then broadcasts them back to the four PEs, in order to process the next time step or the next layer.

In this experimental setup, we note here that the HFINT PE uses MAC operands with 3 exponent bits which was found to yield the best inference accuracy across the ResNet-50, Seq2Seq, and Transformer networks.

The INT and HFINT accelerators were both designed in SystemC with synthesizable and bit-accurate components from the MatchLib~\cite{matchlib} and HLSLibs~\cite{algoc} libraries. Verilog RTL was autogenerated by the Catapult high-level synthesis (HLS) tool with HLS constraints uniformly set with the goal to achieve maximum throughput on the pipelined designs. 

For fair power, performance and area (PPA) comparisons, the two designs employ the same evaluation methodology. Energy and performance results are reported on the post-HLS Verilog netlists by the Catapult tool at 1GHz clock frequency using a commercial 16nm FinFET standard cell library. The simulated workload consists of 100 LSTM time steps with 256 hidden units operating in a weight stationary dataflow. The same process node is also used by Synopsys Design Compiler to extract the area estimates of the accelerators after placement and timing-aware logic synthesis. 

\subsection{Energy, Performance and Area Analyses}
We first look at PPA efficiencies in the PE, which is the computational workhorse of the accelerator. Moreover, we evaluate the effect of increasing throughput via the inner MAC vector size which is also equal to the number of parallel lanes (i.e., vector MAC units). 

\begin{figure}[!t]
\centering
\includegraphics[width=\linewidth]{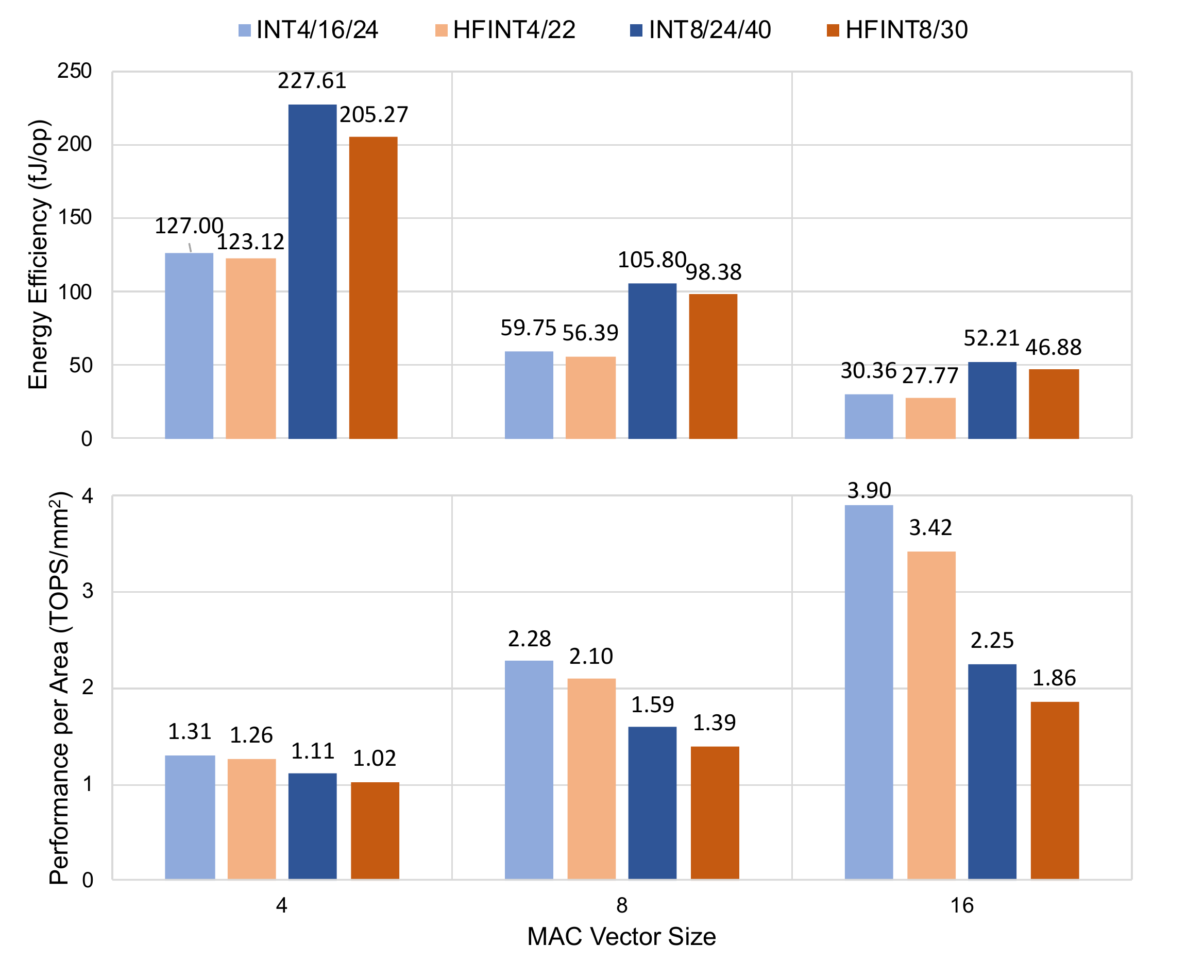}
\caption{Energy efficiency (\textbf{Top}) and throughput per unit area (\textbf{Bottom}) of the INT and HFINT PEs across MAC vector sizes.}
\label{fig:efficiencies}
\vspace{-1.5em}
\end{figure}

Figure~\ref{fig:efficiencies} shows that the HFINT PEs achieve smaller per-operation energy than the INT PEs at either 4-bit or 8-bit MAC operands and across vector sizes. Larger vector size and operand bit width benefits more the HFINT PE than the INT PE in terms of energy efficiency. 
Precisely, from 4-bit operands and vector size of 4 -- to 8-bit operands and vector size of 16, the per-operation energy of the HFINT PE is 0.97$\times$ to 0.90$\times$ that of the INT PE. The smaller per-operation energy of the HFINT PE stems from the fact that its vector MACs contain smaller mantissa multipliers and exponent adders that consume less overall power than full bitwidth multipliers as used in vector MACs of the INT PEs. 

Increasing the vector size is found to improve overall energy efficiency due to higher spatial reuse of the accumulated partial sums. On the other hand, the INT PEs exhibit 1.04$\times$ to 1.21$\times$ higher performance per unit area compared to the HFINT PEs due to the more compact and homogeneous logic in the vector MACs. 

\begin{table}[!t] 
 \caption{PPA results of the 8-bit INT and 8-bit HFINT accelerators}
 \label{tab:ppa_result}
 \vskip 0.05in
  \begin{center}
 \begin{small}
 \begin{sc}
 \begin{adjustbox}{width=\columnwidth,center}
 \centering
\begin{tabular}{llllll}
\toprule
 & Power & Area & Computational Time for \\ & ($mW$) & ($mm^2$) & 100 LSTM Timesteps ($\mu$$s$)\\
\midrule
Int Accelerator & \multirow{2}{*}{61.38}  & \multirow{2}{*}{6.9} & \multirow{2}{*}{81.2}\\
with 4 INT8/24/40 PEs & & & \\ \hline
HFInt Accelerator & \multirow{2}{*}{56.22}  & \multirow{2}{*}{7.9} & \multirow{2}{*}{81.2}\\
with 4 HFINT8/30 PEs & & & \\

\bottomrule
\end{tabular}
\end{adjustbox}
\end{sc}
\end{small}
\end{center}
\vskip -0.1in
\end{table}

Table~\ref{tab:ppa_result} reports the power, area, and compute time of an 8-bit INT and 8-bit HFINT accelerator system with 4 PEs and a global buffer. The PEs here have a MAC vector size of 16 in both systems. The HFINT accelerator reports 0.92$\times$ the power and 1.14$\times$ the area of the integer-based adaptation, confirming the efficiency trends reported in Figure~\ref{fig:efficiencies}. 
Note that both accelerators have the same compute time because the HLS tool generated the same aggregate pipelining result for both designs. 



\section{Conclusion}\label{sec:conclusion}
Fixed-point quantization schemes can be inadequate for networks possessing relatively wide parameter distributions commonly seen in deep sequence transduction models such as the Transformer. This paper presents \NAME, a resilient floating-point based encoding solution that dynamically maximizes and optimally clips its available dynamic range, at a layer granularity, in order to create accurate encodings of neural network parameters from narrow to wide distribution spread. \NAME demonstrates marked robustness at very low precision ($\leq$ 6-bit) on the Transformer, LSTM-based seq2seq, and ResNet-50 networks. This paves the way to higher compute density into reconfigurable architectures -- at a much lower penalty for computational accuracy compared to leading encoding types such as block floating-point, uniform, or non-adaptive float or posit number formats. We also illustrate the algorithm-hardware co-design of \NAME, which allows the extracted $exp_{bias}$ of weights and activations to be stored on allocated registers on-chip in order to perform the adaptive operation of the quantization. The proposed processing elements and accelerators that leverage this mechanism demonstrate per-operation energy that is 0.90$\times$ to 0.97$\times$ that of integer-based adaptations at varying vector sizes and MAC operand bit widths. Altogether, the \NAME algorithm-hardware co-design framework offers a compelling alternative to integer or fixed-point solutions.

\section{Acknowledgments}\label{sec:acknowledgments}
This work was supported by the Application Driving Architectures (ADA) Research Center, a JUMP Center co-sponsored
by SRC and DARPA


\bibliography{example_paper}

\begin{thebibliography}{43}
\providecommand{\natexlab}[1]{#1}
\providecommand{\url}[1]{\texttt{#1}}
\expandafter\ifx\csname urlstyle\endcsname\relax
  \providecommand{\doi}[1]{doi: #1}\else
  \providecommand{\doi}{doi: \begingroup \urlstyle{rm}\Url}\fi

\bibitem[Alom et~al.(2018)Alom, Moody, Maruyama, Essen, and Taha]{qrnn}
Alom, M.~Z., Moody, A.~T., Maruyama, N., Essen, B. C.~V., and Taha, T.~M.
\newblock Effective quantization approaches for recurrent neural networks.
\newblock 2018.
\newblock URL \url{http://arxiv.org/abs/1802.02615}.

\bibitem[Ba et~al.(2016)Ba, Kiros, and Hinton]{layernorm}
Ba, L.~J., Kiros, J.~R., and Hinton, G.~E.
\newblock Layer normalization.
\newblock 2016.
\newblock URL \url{http://arxiv.org/abs/1607.06450}.

\bibitem[Bhandare et~al.(2019)Bhandare, Sripathi, Karkada, Menon, Choi, Datta,
  and Saletore]{int8_transformer}
Bhandare, A., Sripathi, V., Karkada, D., Menon, V., Choi, S., Datta, K., and
  Saletore, V.
\newblock Efficient 8-bit quantization of transformer neural machine language
  translation model.
\newblock 2019.
\newblock URL \url{http://arxiv.org/abs/1906.00532}.

\bibitem[Cai et~al.(2017)Cai, He, Sun, and Vasconcelos]{halfwave}
Cai, Z., He, X., Sun, J., and Vasconcelos, N.
\newblock Deep learning with low precision by half-wave gaussian quantization.
\newblock 2017.
\newblock URL \url{http://arxiv.org/abs/1702.00953}.

\bibitem[Carmichael et~al.(2018)Carmichael, Langroudi, Khazanov, Lillie,
  Gustafson, and Kudithipudi]{positron}
Carmichael, Z., Langroudi, S. H.~F., Khazanov, C., Lillie, J., Gustafson,
  J.~L., and Kudithipudi, D.
\newblock Deep positron: {A} deep neural network using the posit number system.
\newblock 2018.
\newblock URL \url{http://arxiv.org/abs/1812.01762}.

\bibitem[Chiu et~al.(2017)Chiu, Sainath, Wu, Prabhavalkar, Nguyen, Chen,
  Kannan, Weiss, Rao, Gonina, Jaitly, Li, Chorowski, and Bacchiani]{sotaspeech}
Chiu, C., Sainath, T.~N., Wu, Y., Prabhavalkar, R., Nguyen, P., Chen, Z.,
  Kannan, A., Weiss, R.~J., Rao, K., Gonina, K., Jaitly, N., Li, B., Chorowski,
  J., and Bacchiani, M.
\newblock State-of-the-art speech recognition with sequence-to-sequence models.
\newblock 2017.
\newblock URL \url{http://arxiv.org/abs/1712.01769}.

\bibitem[Choi et~al.(2019)Choi, Venkataramani, Srinivasan, Gopalakrishnan,
  Wang, and Chuang]{Choi2019ACCURATEAE}
Choi, J., Venkataramani, S., Srinivasan, V., Gopalakrishnan, K., Wang, Z., and
  Chuang, P.
\newblock Accurate and efficient 2-bit quantized neural networks.
\newblock 2019.

\bibitem[Chorowski et~al.(2015)Chorowski, Bahdanau, Serdyuk, Cho, and
  Bengio]{seq2seq}
Chorowski, J., Bahdanau, D., Serdyuk, D., Cho, K., and Bengio, Y.
\newblock Attention-based models for speech recognition.
\newblock In \emph{Advances in Neural Information Processing Systems 28: Annual
  Conference on Neural Information Processing Systems 2015, December 7-12,
  2015, Montreal, Quebec, Canada}, pp.\  577--585, 2015.

\bibitem[Courbariaux \& Bengio(2016)Courbariaux and Bengio]{binarynet}
Courbariaux, M. and Bengio, Y.
\newblock Binarynet: Training deep neural networks with weights and activations
  constrained to +1 or -1.
\newblock 2016.
\newblock URL \url{http://arxiv.org/abs/1602.02830}.

\bibitem[Courbariaux et~al.(2015)Courbariaux, Bengio, and
  David]{CourbariauxBD15}
Courbariaux, M., Bengio, Y., and David, J.
\newblock Binaryconnect: Training deep neural networks with binary weights
  during propagations.
\newblock 2015.
\newblock URL \url{http://arxiv.org/abs/1511.00363}.

\bibitem[Drumond et~al.(2018)Drumond, Lin, Jaggi, and Falsafi]{bfp2018}
Drumond, M., Lin, T., Jaggi, M., and Falsafi, B.
\newblock End-to-end {DNN} training with block floating point arithmetic.
\newblock 2018.
\newblock URL \url{http://arxiv.org/abs/1804.01526}.

\bibitem[Fowers et~al.(2018)Fowers, Ovtcharov, Papamichael, Massengill, Liu,
  Lo, Alkalay, Haselman, Adams, Ghandi, et~al.]{brainwave}
Fowers, J., Ovtcharov, K., Papamichael, M., Massengill, T., Liu, M., Lo, D.,
  Alkalay, S., Haselman, M., Adams, L., Ghandi, M., et~al.
\newblock A configurable cloud-scale dnn processor for real-time ai.
\newblock In \emph{Proceedings of the 45th Annual International Symposium on
  Computer Architecture}. pages 1-14. IEEE Press, 2018.

\bibitem[Gupta et~al.(2015)Gupta, Agrawal, Gopalakrishnan, and
  Narayanan]{ibm2015}
Gupta, S., Agrawal, A., Gopalakrishnan, K., and Narayanan, P.
\newblock Deep learning with limited numerical precision.
\newblock 2015.
\newblock URL \url{http://arxiv.org/abs/1502.02551}.

\bibitem[Gustafson \& Yonemoto(2017)Gustafson and Yonemoto]{posit2017}
Gustafson and Yonemoto.
\newblock Beating floating point at its own game: Posit arithmetic.
\newblock June 2017.
\newblock ISSN 2409-6008.
\newblock \doi{10.14529/jsfi170206}.
\newblock URL \url{https://doi.org/10.14529/jsfi170206}.

\bibitem[Han et~al.(2015)Han, Mao, and Dally]{deepcompression}
Han, S., Mao, H., and Dally, W.~J.
\newblock Deep compression: Compressing deep neural network with pruning,
  trained quantization and huffman coding.
\newblock 2015.

\bibitem[{He} et~al.(2016){He}, {Zhang}, {Ren}, and {Sun}]{resnet}
{He}, K., {Zhang}, X., {Ren}, S., and {Sun}, J.
\newblock Deep residual learning for image recognition.
\newblock In \emph{2016 IEEE Conference on Computer Vision and Pattern
  Recognition (CVPR)}, pp.\  770--778, June 2016.
\newblock \doi{10.1109/CVPR.2016.90}.

\bibitem[HLSLibs()]{algoc}
HLSLibs.
\newblock Open-source high-level synthesis ip libraries.
\newblock Technical report.
\newblock URL \url{https://github.com/hlslibs}.

\bibitem[{Hwang} \& {Sung}(2014){Hwang} and {Sung}]{sips2014}
{Hwang}, K. and {Sung}, W.
\newblock Fixed-point feedforward deep neural network design using weights +1,
  0, and −1.
\newblock In \emph{2014 IEEE Workshop on Signal Processing Systems (SiPS)},
  pp.\  1--6, Oct 2014.
\newblock \doi{10.1109/SiPS.2014.6986082}.

\bibitem[Jacob et~al.(2017)Jacob, Kligys, Chen, Zhu, Tang, Howard, Adam, and
  Kalenichenko]{googleint}
Jacob, B., Kligys, S., Chen, B., Zhu, M., Tang, M., Howard, A.~G., Adam, H.,
  and Kalenichenko, D.
\newblock Quantization and training of neural networks for efficient
  integer-arithmetic-only inference.
\newblock 2017.
\newblock URL \url{http://arxiv.org/abs/1712.05877}.

\bibitem[Johnson(2018)]{rethinkfloat}
Johnson, J.
\newblock Rethinking floating point for deep learning.
\newblock 2018.
\newblock URL \url{http://arxiv.org/abs/1811.01721}.

\bibitem[{Jouppi} et~al.(2017){Jouppi}, {Young}, {Patil}, {Patterson},
  {Agrawal}, {Bajwa}, {Bates}, {Bhatia}, {Boden}, {Borchers}, {Boyle},
  {Cantin}, {Chao}, {Clark}, {Coriell}, {Daley}, {Dau}, {Dean}, {Gelb},
  {Ghaemmaghami}, {Gottipati}, {Gulland}, {Hagmann}, {Ho}, {Hogberg}, {Hu},
  {Hundt}, {Hurt}, {Ibarz}, {Jaffey}, {Jaworski}, {Kaplan}, {Khaitan},
  {Killebrew}, {Koch}, {Kumar}, {Lacy}, {Laudon}, {Law}, {Le}, {Leary}, {Liu},
  {Lucke}, {Lundin}, {MacKean}, {Maggiore}, {Mahony}, {Miller}, {Nagarajan},
  {Narayanaswami}, {Ni}, {Nix}, {Norrie}, {Omernick}, {Penukonda}, {Phelps},
  {Ross}, {Ross}, {Salek}, {Samadiani}, {Severn}, {Sizikov}, {Snelham},
  {Souter}, {Steinberg}, {Swing}, {Tan}, {Thorson}, {Tian}, {Toma}, {Tuttle},
  {Vasudevan}, {Walter}, {Wang}, {Wilcox}, and {Yoon}]{Jouppi2017}
{Jouppi}, N.~P., {Young}, C., {Patil}, N., {Patterson}, D., {Agrawal}, G.,
  {Bajwa}, R., {Bates}, S., {Bhatia}, S., {Boden}, N., {Borchers}, A., {Boyle},
  R., {Cantin}, P., {Chao}, C., {Clark}, C., {Coriell}, J., {Daley}, M., {Dau},
  M., {Dean}, J., {Gelb}, B., {Ghaemmaghami}, T.~V., {Gottipati}, R.,
  {Gulland}, W., {Hagmann}, R., {Ho}, C.~R., {Hogberg}, D., {Hu}, J., {Hundt},
  R., {Hurt}, D., {Ibarz}, J., {Jaffey}, A., {Jaworski}, A., {Kaplan}, A.,
  {Khaitan}, H., {Killebrew}, D., {Koch}, A., {Kumar}, N., {Lacy}, S.,
  {Laudon}, J., {Law}, J., {Le}, D., {Leary}, C., {Liu}, Z., {Lucke}, K.,
  {Lundin}, A., {MacKean}, G., {Maggiore}, A., {Mahony}, M., {Miller}, K.,
  {Nagarajan}, R., {Narayanaswami}, R., {Ni}, R., {Nix}, K., {Norrie}, T.,
  {Omernick}, M., {Penukonda}, N., {Phelps}, A., {Ross}, J., {Ross}, M.,
  {Salek}, A., {Samadiani}, E., {Severn}, C., {Sizikov}, G., {Snelham}, M.,
  {Souter}, J., {Steinberg}, D., {Swing}, A., {Tan}, M., {Thorson}, G., {Tian},
  B., {Toma}, H., {Tuttle}, E., {Vasudevan}, V., {Walter}, R., {Wang}, W.,
  {Wilcox}, E., and {Yoon}, D.~H.
\newblock In-datacenter performance analysis of a tensor processing unit.
\newblock In \emph{2017 ACM/IEEE 44th Annual International Symposium on
  Computer Architecture (ISCA)}, pp.\  1--12, June 2017.
\newblock \doi{10.1145/3079856.3080246}.

\bibitem[Khailany et~al.(2018)Khailany, Khmer, Venkatesan, Clemons, Emer,
  Fojtik, Klinefelter, Pellauer, Pinckney, Shao, Srinath, Torng, Xi, Zhang, and
  Zimmer]{matchlib}
Khailany, B., Khmer, E., Venkatesan, R., Clemons, J., Emer, J.~S., Fojtik, M.,
  Klinefelter, A., Pellauer, M., Pinckney, N., Shao, Y.~S., Srinath, S., Torng,
  C., Xi, S.~L., Zhang, Y., and Zimmer, B.
\newblock A modular digital vlsi flow for high-productivity soc design.
\newblock In \emph{Proceedings of the 55th Annual Design Automation
  Conference}, DAC '18, pp.\  72:1--72:6, New York, NY, USA, 2018. ACM.
\newblock ISBN 978-1-4503-5700-5.
\newblock \doi{10.1145/3195970.3199846}.
\newblock URL \url{http://doi.acm.org/10.1145/3195970.3199846}.

\bibitem[Klein et~al.(2017)Klein, Kim, Deng, Senellart, and Rush]{onmt}
Klein, G., Kim, Y., Deng, Y., Senellart, J., and Rush, A.~M.
\newblock Opennmt: Open-source toolkit for neural machine translation.
\newblock 2017.
\newblock URL \url{http://arxiv.org/abs/1701.02810}.

\bibitem[K{\"{o}}ster et~al.(2017)K{\"{o}}ster, Webb, Wang, Nassar, Bansal,
  Constable, Elibol, Hall, Hornof, Khosrowshahi, Kloss, Pai, and
  Rao]{flexpoint}
K{\"{o}}ster, U., Webb, T., Wang, X., Nassar, M., Bansal, A.~K., Constable, W.,
  Elibol, O., Hall, S., Hornof, L., Khosrowshahi, A., Kloss, C., Pai, R.~J.,
  and Rao, N.
\newblock Flexpoint: An adaptive numerical format for efficient training of
  deep neural networks.
\newblock 2017.
\newblock URL \url{http://arxiv.org/abs/1711.02213}.

\bibitem[Krizhevsky et~al.(2012)Krizhevsky, Sutskever, and Hinton]{CNN12}
Krizhevsky, A., Sutskever, I., and Hinton, G.~E.
\newblock Imagenet classification with deep convolutional neural networks.
\newblock In \emph{Proceedings of the 25th International Conference on Neural
  Information Processing Systems - Volume 1}, NIPS'12, pp.\  1097--1105, USA,
  2012. Curran Associates Inc.
\newblock URL \url{http://dl.acm.org/citation.cfm?id=2999134.2999257}.

\bibitem[{Lee} et~al.(2017){Lee}, {Miyashita}, {Chai}, {Murmann}, and
  {Wong}]{lognet}
{Lee}, E.~H., {Miyashita}, D., {Chai}, E., {Murmann}, B., and {Wong}, S.~S.
\newblock Lognet: Energy-efficient neural networks using logarithmic
  computation.
\newblock In \emph{2017 IEEE International Conference on Acoustics, Speech and
  Signal Processing (ICASSP)}, pp.\  5900--5904, March 2017.
\newblock \doi{10.1109/ICASSP.2017.7953288}.

\bibitem[Lin et~al.(2015)Lin, Talathi, and Annapureddy]{qualcomm_fixed_point}
Lin, D.~D., Talathi, S.~S., and Annapureddy, V.~S.
\newblock Fixed point quantization of deep convolutional networks.
\newblock 2015.
\newblock URL \url{http://arxiv.org/abs/1511.06393}.

\bibitem[Migacz(2017)]{migacz2017}
Migacz, S.
\newblock 8-bit inference with tensorrt.
\newblock In \emph{NVIDIA GPU Technology Conference}, 2017.
\newblock URL
  \url{http://on-demand.gputechconf.com/gtc/2017/presentation/s7310-8-bit-inference-with-tensorrt.pdf}.

\bibitem[Mishra et~al.(2017)Mishra, Nurvitadhi, Cook, and Marr]{wrpn}
Mishra, A.~K., Nurvitadhi, E., Cook, J.~J., and Marr, D.
\newblock {WRPN:} wide reduced-precision networks.
\newblock 2017.
\newblock URL \url{http://arxiv.org/abs/1709.01134}.

\bibitem[Miyashita et~al.(2016)Miyashita, Lee, and Murmann]{logcnn}
Miyashita, D., Lee, E.~H., and Murmann, B.
\newblock Convolutional neural networks using logarithmic data representation.
\newblock 2016.
\newblock URL \url{http://arxiv.org/abs/1603.01025}.

\bibitem[Noh et~al.(2017)Noh, You, Mun, and Han]{deepnoise}
Noh, H., You, T., Mun, J., and Han, B.
\newblock Regularizing deep neural networks by noise: Its interpretation and
  optimization.
\newblock In \emph{Advances in Neural Information Processing Systems 30: Annual
  Conference on Neural Information Processing Systems 2017, 4-9 December 2017,
  Long Beach, CA, {USA}}, pp.\  5109--5118, 2017.

\bibitem[{Park} et~al.(2017){Park}, {Ahn}, and {Yoo}]{wen}
{Park}, E., {Ahn}, J., and {Yoo}, S.
\newblock Weighted-entropy-based quantization for deep neural networks.
\newblock In \emph{2017 IEEE Conference on Computer Vision and Pattern
  Recognition (CVPR)}, pp.\  7197--7205, July 2017.
\newblock \doi{10.1109/CVPR.2017.761}.

\bibitem[{Park} et~al.(2018){Park}, {Kim}, and {Yoo}]{Park2018}
{Park}, E., {Kim}, D., and {Yoo}, S.
\newblock Energy-efficient neural network accelerator based on outlier-aware
  low-precision computation.
\newblock In \emph{2018 ACM/IEEE 45th Annual International Symposium on
  Computer Architecture (ISCA)}, pp.\  688--698, June 2018.
\newblock \doi{10.1109/ISCA.2018.00063}.

\bibitem[Pytorch()]{pytorch_imagenet}
Pytorch.
\newblock Imagenet training in pytorch.
\newblock Technical report.
\newblock URL \url{https://github.com/pytorch/examples/tree/master/imagenet}.

\bibitem[{Reagen} et~al.(2016){Reagen}, {Whatmough}, {Adolf}, {Rama}, {Lee},
  {Lee}, {Hernández-Lobato}, {Wei}, and {Brooks}]{minerva}
{Reagen}, B., {Whatmough}, P., {Adolf}, R., {Rama}, S., {Lee}, H., {Lee},
  S.~K., {Hernández-Lobato}, J.~M., {Wei}, G., and {Brooks}, D.
\newblock Minerva: Enabling low-power, highly-accurate deep neural network
  accelerators.
\newblock In \emph{2016 ACM/IEEE 43rd Annual International Symposium on
  Computer Architecture (ISCA)}, pp.\  267--278, June 2016.
\newblock \doi{10.1109/ISCA.2016.32}.

\bibitem[Salimans \& Kingma(2016)Salimans and Kingma]{weightnorm}
Salimans, T. and Kingma, D.~P.
\newblock Weight normalization: {A} simple reparameterization to accelerate
  training of deep neural networks.
\newblock In \emph{Advances in Neural Information Processing Systems 29: Annual
  Conference on Neural Information Processing Systems 2016, December 5-10,
  2016, Barcelona, Spain}, pp.\  901, 2016.
\newblock URL
  \url{http://papers.nips.cc/paper/6114-weight-normalization-a-simple-reparameterization-to-accelerate-training-of-deep-neural-networks}.

\bibitem[Vaswani et~al.(2017)Vaswani, Shazeer, Parmar, Uszkoreit, Jones, Gomez,
  Kaiser, and Polosukhin]{attention}
Vaswani, A., Shazeer, N., Parmar, N., Uszkoreit, J., Jones, L., Gomez, A.~N.,
  Kaiser, L., and Polosukhin, I.
\newblock Attention is all you need.
\newblock 2017.
\newblock URL \url{http://arxiv.org/abs/1706.03762}.

\bibitem[Vogel et~al.(2018)Vogel, Liang, Guntoro, Stechele, and
  Ascheid]{logbase}
Vogel, S., Liang, M., Guntoro, A., Stechele, W., and Ascheid, G.
\newblock Efficient hardware acceleration of cnns using logarithmic data
  representation with arbitrary log-base.
\newblock In \emph{Proceedings of the International Conference on
  Computer-Aided Design}, ICCAD '18, pp.\  9:1--9:8, New York, NY, USA, 2018.
  ACM.
\newblock ISBN 978-1-4503-5950-4.
\newblock \doi{10.1145/3240765.3240803}.
\newblock URL \url{http://doi.acm.org/10.1145/3240765.3240803}.

\bibitem[Wu et~al.(2015)Wu, Leng, Wang, Hu, and Cheng]{conv_mobiles}
Wu, J., Leng, C., Wang, Y., Hu, Q., and Cheng, J.
\newblock Quantized convolutional neural networks for mobile devices.
\newblock 2015.
\newblock URL \url{http://arxiv.org/abs/1512.06473}.

\bibitem[Wu et~al.(2016)Wu, Schuster, Chen, Le, Norouzi, Macherey, Krikun, Cao,
  Gao, Macherey, Klingner, Shah, Johnson, Liu, Kaiser, Gouws, Kato, Kudo,
  Kazawa, Stevens, Kurian, Patil, Wang, Young, Smith, Riesa, Rudnick, Vinyals,
  Corrado, Hughes, and Dean]{googlenmt}
Wu, Y., Schuster, M., Chen, Z., Le, Q.~V., Norouzi, M., Macherey, W., Krikun,
  M., Cao, Y., Gao, Q., Macherey, K., Klingner, J., Shah, A., Johnson, M., Liu,
  X., Kaiser, L., Gouws, S., Kato, Y., Kudo, T., Kazawa, H., Stevens, K.,
  Kurian, G., Patil, N., Wang, W., Young, C., Smith, J., Riesa, J., Rudnick,
  A., Vinyals, O., Corrado, G., Hughes, M., and Dean, J.
\newblock Google's neural machine translation system: Bridging the gap between
  human and machine translation.
\newblock 2016.
\newblock URL \url{http://arxiv.org/abs/1609.08144}.

\bibitem[Zhang et~al.(2018)Zhang, Yang, Ye, and Hua]{lqnets}
Zhang, D., Yang, J., Ye, D., and Hua, G.
\newblock Lq-nets: Learned quantization for highly accurate and compact deep
  neural networks.
\newblock 2018.
\newblock URL \url{http://arxiv.org/abs/1807.10029}.

\bibitem[Zhou et~al.(2016)Zhou, Ni, Zhou, Wen, Wu, and Zou]{dorefanet}
Zhou, S., Ni, Z., Zhou, X., Wen, H., Wu, Y., and Zou, Y.
\newblock Dorefa-net: Training low bitwidth convolutional neural networks with
  low bitwidth gradients.
\newblock 2016.
\newblock URL \url{http://arxiv.org/abs/1606.06160}.

\bibitem[Zhu et~al.(2016)Zhu, Han, Mao, and Dally]{ternary}
Zhu, C., Han, S., Mao, H., and Dally, W.~J.
\newblock Trained ternary quantization.
\newblock 2016.
\newblock URL \url{http://arxiv.org/abs/1612.01064}.

\end{thebibliography}
\bibliographystyle{sysml2019}

\appendix
%


\end{document}